\def\year{2018}\relax
\documentclass[letterpaper]{article} 
\usepackage{aaai18}  
\usepackage{times}  
\usepackage{helvet}  
\usepackage{courier}  
\usepackage{url}  
\usepackage{graphicx}  

\usepackage{amsfonts}
\usepackage{booktabs}
\usepackage{multirow}
\usepackage{floatrow}
\usepackage{epstopdf}
\usepackage{array}
\usepackage{amsmath}
\usepackage[colorlinks,linkcolor=red]{hyperref}
\begin{document}
%
\title{M2Det: A Single-Shot Object Detector based on Multi-Level Feature Pyramid Network}

\author{Qijie~Zhao$^1$, Tao~Sheng$^1$,Yongtao~Wang$^1$\thanks{Corresponding author.}, Zhi~Tang$^1$,  Ying Chen$^2$, Ling~Cai$^2$ and Haibin Ling$^3$
\\
\normalsize
$^1$Institute of Computer Science and Technology, Peking University, Beijing, P.R. China
\\
\normalsize
$^2$AI Labs, DAMO Academy, Alibaba Group
\\
\normalsize
$^3$Computer and Information Sciences Department, Temple University
\\
\tt\small
\{zhaoqijie, shengtao, wyt, tangzhi\}@pku.edu.cn,\\ 
\tt\small
\{cailing.cl, chenying.ailab\}@alibaba-inc.com, \{hbling\}@temple.edu
}

\maketitle
\begin{abstract}
Feature pyramids are widely exploited by both the state-of-the-art one-stage object detectors (\textit{e.g.}, DSSD, RetinaNet, RefineDet) and the two-stage object detectors (\textit{e.g.}, Mask R-CNN, DetNet) to alleviate the problem arising from scale variation across object instances. Although these object detectors with feature pyramids achieve encouraging results, they have some limitations due to that they only simply construct the feature pyramid according to the inherent multi-scale, pyramidal architecture of the backbones which are originally designed for object classification task. Newly, in this work, we present Multi-Level Feature Pyramid Network (MLFPN) to construct more effective feature pyramids for detecting objects of different scales. First, we fuse multi-level features (\textit{i.e.} multiple layers) extracted by backbone as the base feature. Second, we feed the base feature into a block of alternating joint Thinned U-shape Modules and Feature Fusion Modules and exploit the decoder layers of each U-shape module as the features for detecting objects. Finally, we gather up the decoder layers with equivalent scales (sizes) to construct a feature pyramid for object detection, in which every feature map consists of the layers (features) from multiple levels. To evaluate the effectiveness of the proposed MLFPN, we design and train a powerful end-to-end one-stage object detector we call M2Det by integrating it into the architecture of SSD, and achieve better detection performance than state-of-the-art one-stage detectors. Specifically, on MS-COCO benchmark, M2Det achieves AP of 41.0 at speed of 11.8 FPS with single-scale inference strategy and AP of 44.2 with multi-scale inference strategy, which are the new state-of-the-art results among one-stage detectors. The code will be made available on \url{https://github.com/qijiezhao/M2Det}.
\end{abstract}

\section{Introduction}
Scale variation across object instances is one of the major challenges for the object detection task \cite{LinDGHHB17,HeZR015,abs-1711-08189}, and usually there are two strategies to solve the problem arising from this challenge. The first one is to detect objects in an image pyramid (\textit{i.e.} a series of resized copies of the input image) \cite{abs-1711-08189}, which can only be exploited at the testing time. Obviously, this solution will greatly increase memory and computational complexity, thus the efficiency of such object detectors drop dramatically. The second one is to detect objects in a feature pyramid extracted from the input image \cite{LiuAESRFB16,LinDGHHB17}, which can be exploited at both training and testing phases. Compared with the first solution that uses an image pyramid, it has less memory and computational cost. Moreover, the feature pyramid constructing module can be easily integrated into the state-of-the-art deep neural networks based detectors, yielding an end-to-end solution.

\begin{figure}[t]
\centering
\includegraphics[scale=0.39]{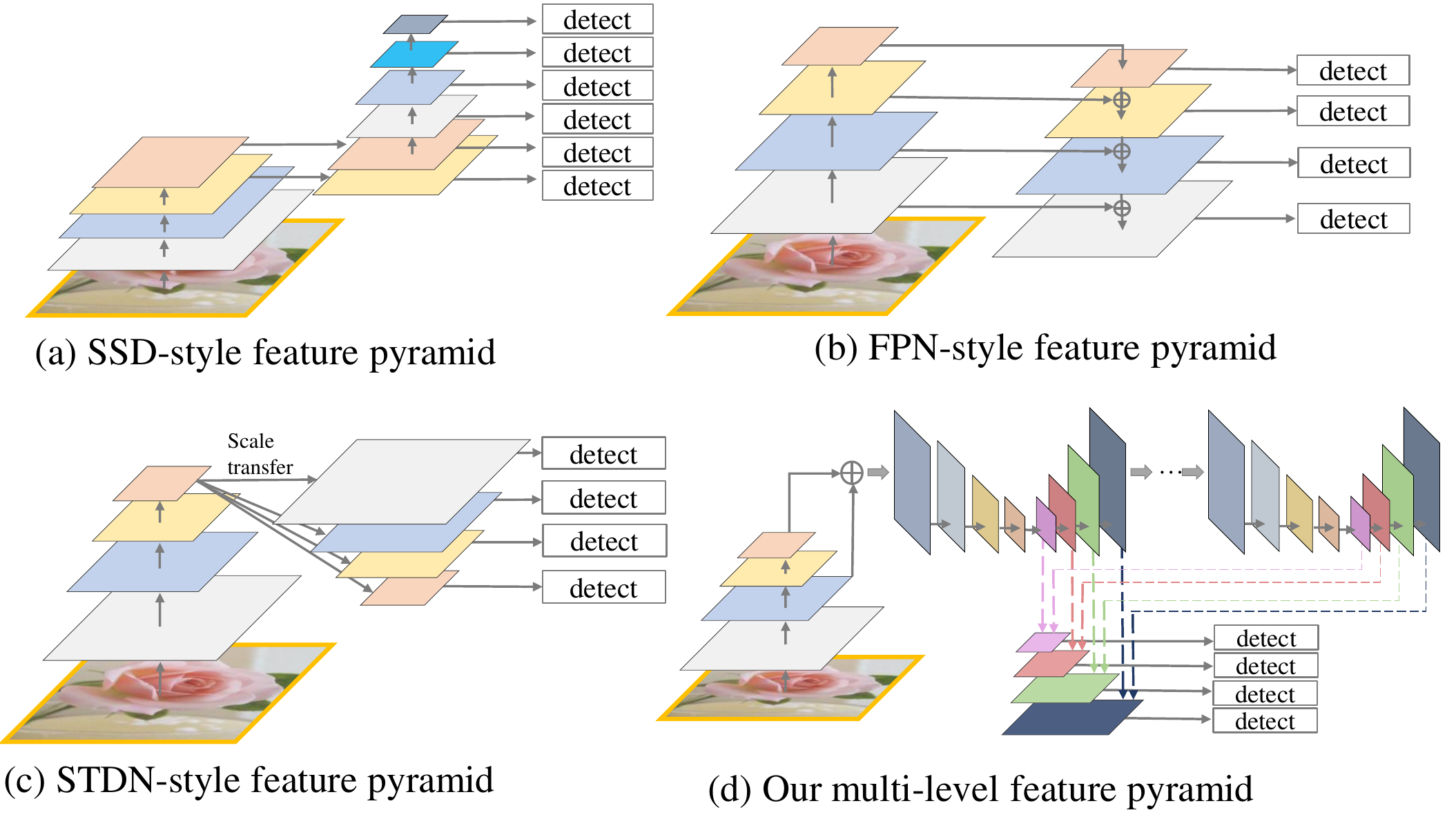}
\caption{Illustrations of four kinds of feature pyramids.}
\label{fig:pyram}
\end{figure}

Although the object detectors with feature pyramids \cite{LiuAESRFB16,LinDGHHB17,LinGGHD17,HeGDG17} achieve encouraging results, they still have some limitations due to that they simply construct the feature pyramid according to the inherent multi-scale, pyramidal architecture of the backbones which are actually designed for object classification task. For example, as illustrated in Fig. \ref{fig:pyram}, SSD \cite{LiuAESRFB16} directly and independently uses two layers of the backbone (\textit{i.e.} VGG16) and four extra layers obtained by stride 2 convolution to construct the feature pyramid; STDN \cite{zhou2018scale} only uses the last dense block of DenseNet \cite{huang2017densely} to construct feature pyramid by pooling and scale-transfer operations; FPN \cite{LinDGHHB17} constructs the feature pyramid by fusing the deep and shallow layers in a top-down manner. Generally speaking, the above-mentioned methods have the two following limitations. First, feature maps in the pyramid are not representative enough for the object detection task, since they are simply constructed from the layers (features) of the backbone designed for object classification task. Second, each feature map in the pyramid (used for detecting objects in a specific range of size) is mainly or even solely constructed from single-level layers of the backbone, that is, it mainly or only contains single-level information. In general, high-level features in the deeper layers are more discriminative for classification subtask while low-level features in the shallower layers can be helpful for object location regression sub-task. Moreover, low-level features are more suitable to characterize objects with simple appearances while high-level features are appropriate for objects with complex appearances. In practice, the appearances of the object instances with similar size can be quite different. For example, a traffic light and a faraway person may have comparable size, and the appearance of the person is much more complex. Hence, each feature map (used for detecting objects in a specific range of size) in the pyramid mainly or only consists of single-level features will result in suboptimal detection performance.
\begin{figure*}[t]
\centering
\includegraphics[width=16.5cm]{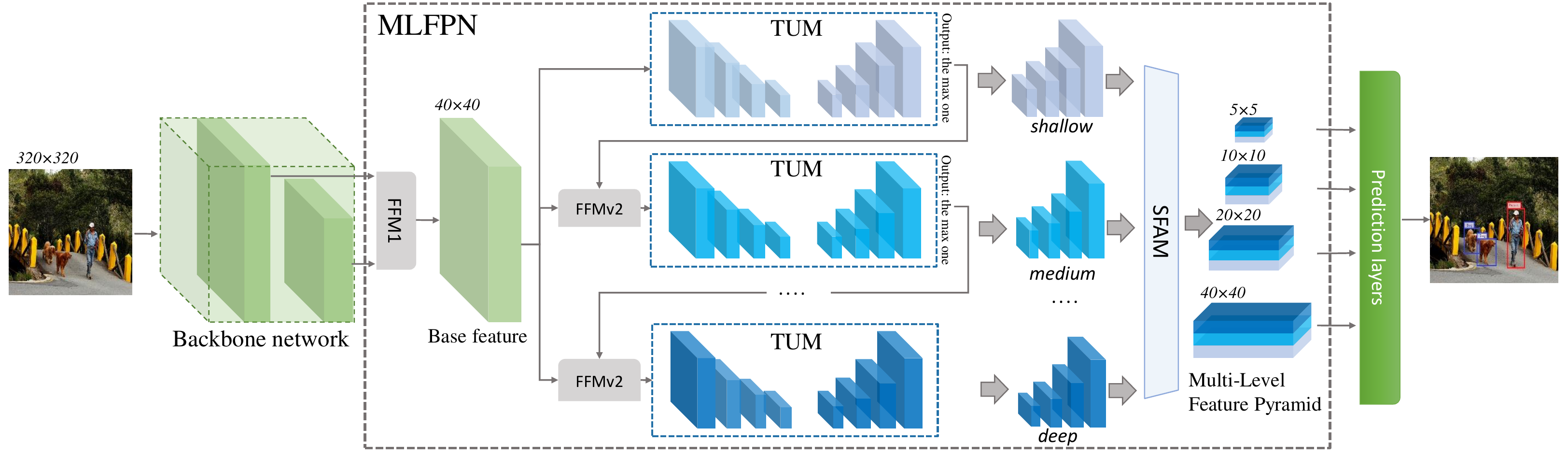}
\caption{An overview of the proposed M2Det($320\times320$). M2Det utilizes the backbone and the Multi-level Feature Pyramid Network (MLFPN) to extract features from the input image, and then produces dense bounding boxes and category scores. In MLFPN, FFMv1 fuses feature maps of the backbone to generate the base feature. Each TUM generates a group of multi-scale features, and then the alternating joint TUMs and FFMv2s extract multi-level multi-scale features. Finally, SFAM aggregates the features into a multi-level feature pyramid. In practice, we use 6 scales and 8 levels.}
\label{fig:pipeline}
\end{figure*}

\begin{figure*}
\centering
\includegraphics[width=14cm]{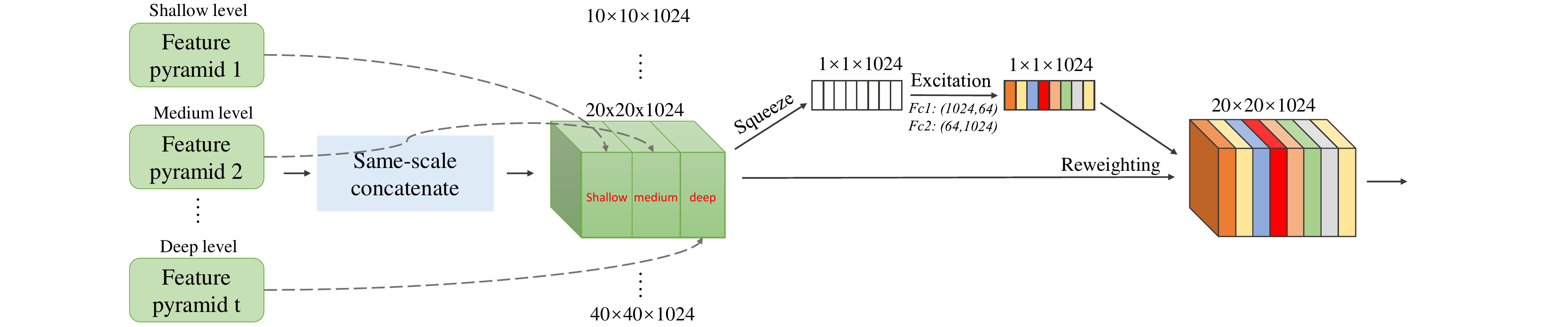}
\caption{Illustration of Scale-wise Feature Aggregation Module. The first stage of SFAM is to concatenate features with equivalent scales along channel dimension. Then the second stage uses SE attention to aggregate features in an adaptive way.}
\label{fig:attention}
\end{figure*}

The goal of this paper is to construct a more effective feature pyramid for detecting objects of different scales, while avoid the limitations of the existing methods as above mentioned. As shown in Fig. \ref{fig:pipeline}, to achieve this goal, we first fuse multi-level features (\textit{i.e.} multiple layers) extracted by backbone as base feature, and then feed it into a block of alternating joint Thinned U-shape Modules(TUM) and Feature Fusion Modules(FFM) to extract more representative, multi-level multi-scale features. 
It is worth noting that, decoder layers in each U-shape Module share a similar depth. Finally, we gather up the feature maps with equivalent scales to construct the final feature pyramid for object detection. Obviously, decoder layers that form the final feature pyramid are much deeper than the layers in the backbone, namely, they are more representative. Moreover, each feature map in the final feature pyramid consists of the decoder layers from multiple levels. Hence, we call our feature pyramid block Multi-Level Feature Pyramid Network (MLFPN).

To evaluate the effectiveness of the proposed MLFPN, we design and train a powerful end-to-end one-stage object detector we call M2Det (according to that it is built upon multi-level and multi-scale features) by integrating MLFPN into the architecture of SSD \cite{LiuAESRFB16}. M2Det achieves the new state-of-the-art result (\textit{i.e.} AP of 41.0 at speed of 11.8 FPS with single-scale inference strategy and AP of 44.2 with multi-scale inference strategy), outperforming the one-stage detectors on MS-COCO \cite{LinMBHPRDZ14} benchmark.

\section{Related Work}
Researchers have put plenty of efforts into improving the detection accuracy of objects with various scales -- no matter what kind of detector it is, either an one-stage detector or a two-stage one. To the best of our knowledge, there are mainly two strategies to tackle this scale-variation problem.

The first one is \textbf{featurizing image pyramids} (\textit{i.e.} a series of resized copies of the input image) to produce semantically representative multi-scale features. Features from images of different scales yield predictions separately and these predictions work together to give the final prediction. In terms of recognition accuracy and localization precision, features from various-sized images do surpass features that are based merely on single-scale images. Methods such as \cite{ShrivastavaGG16} and SNIP \cite{abs-1711-08189} employed this tactic. Despite the performance gain, such a strategy could be costly time-wise and memory-wise, which forbid its application in real-time tasks. Considering this major drawback, methods such as SNIP \cite{abs-1711-08189} can choose to only employ featurized image pyramids during the test phase as a fallback, whereas other methods including Fast R-CNN \cite{Girshick15} and Faster R-CNN \cite{RenHGS15} chose not to use this strategy by default.

The second one is detecting objects in the \textbf{feature pyramid} extracted from inherent layers within the network while merely taking a single-scale image. This strategy demands significantly less additional memory and computational cost than the first one, enabling deployment during both the training and test phases in real-time networks. Moreover, the feature pyramid constructing module can be easily revised and fit into state-of-the-art deep neural networks based detectors. MS-CNN \cite{CaiFFV16}, SSD \cite{LiuAESRFB16}, DSSD \cite{FuLRTB17}, FPN \cite{LinDGHHB17}, YOLOv3 \cite{yolov3}, RetinaNet \cite{LinGGHD17}, and RefineDet \cite{abs-1711-06897} adopted this tactic in different ways.

To the best of our knowledge, MS-CNN \cite{CaiFFV16} proposed two sub-networks and first incorporated multi-scale features into deep convolutional neural networks for object detection. The proposal sub-net exploited feature maps of several resolutions to detect multi-scale objects in an image. SSD \cite{LiuAESRFB16} exploited feature maps from the later layers of VGG16 base-net and extra feature layers for predictions at multiple scales. FPN \cite{LinDGHHB17} utilized lateral connections and a top-down pathway to produce a feature pyramid and achieved more powerful representations. 
DSSD \cite{FuLRTB17} implemented deconvolution layers for aggregating context and enhancing the high-level semantics for shallow features. RefineDet \cite{abs-1711-06897} adopted two-step cascade regression, which achieves a remarkable progress on accuracy while keeping the efficiency of SSD.

\section{Proposed Method}
The overall architecture of M2Det is shown in Fig. \ref{fig:pipeline}. M2Det uses the backbone and the Multi-Level Feature Pyramid Network (MLFPN) to extract  features from the input image, and then similar to SSD, produces dense bounding boxes and category scores based on the learned features, followed by the non-maximum suppression (NMS) operation to produce the final results. MLFPN consists of three modules, \textit{i.e.} Feature Fusion Module (FFM), Thinned U-shape Module (TUM) and Scale-wise Feature Aggregation Module (SFAM). FFMv1 enriches semantic information into base features by fusing feature maps of the backbone. Each TUM generates a group of multi-scale features, and then the alternating joint TUMs and FFMv2s extract multi-level multi-scale features. In addition, SFAM aggregates the features into the multi-level feature pyramid through a scale-wise feature concatenation operation and an adaptive attention mechanism. More details about the three core modules and network configurations in M2Det are introduced in the following.

\begin{figure}[!t]
	\centering
	\includegraphics[scale=0.32]{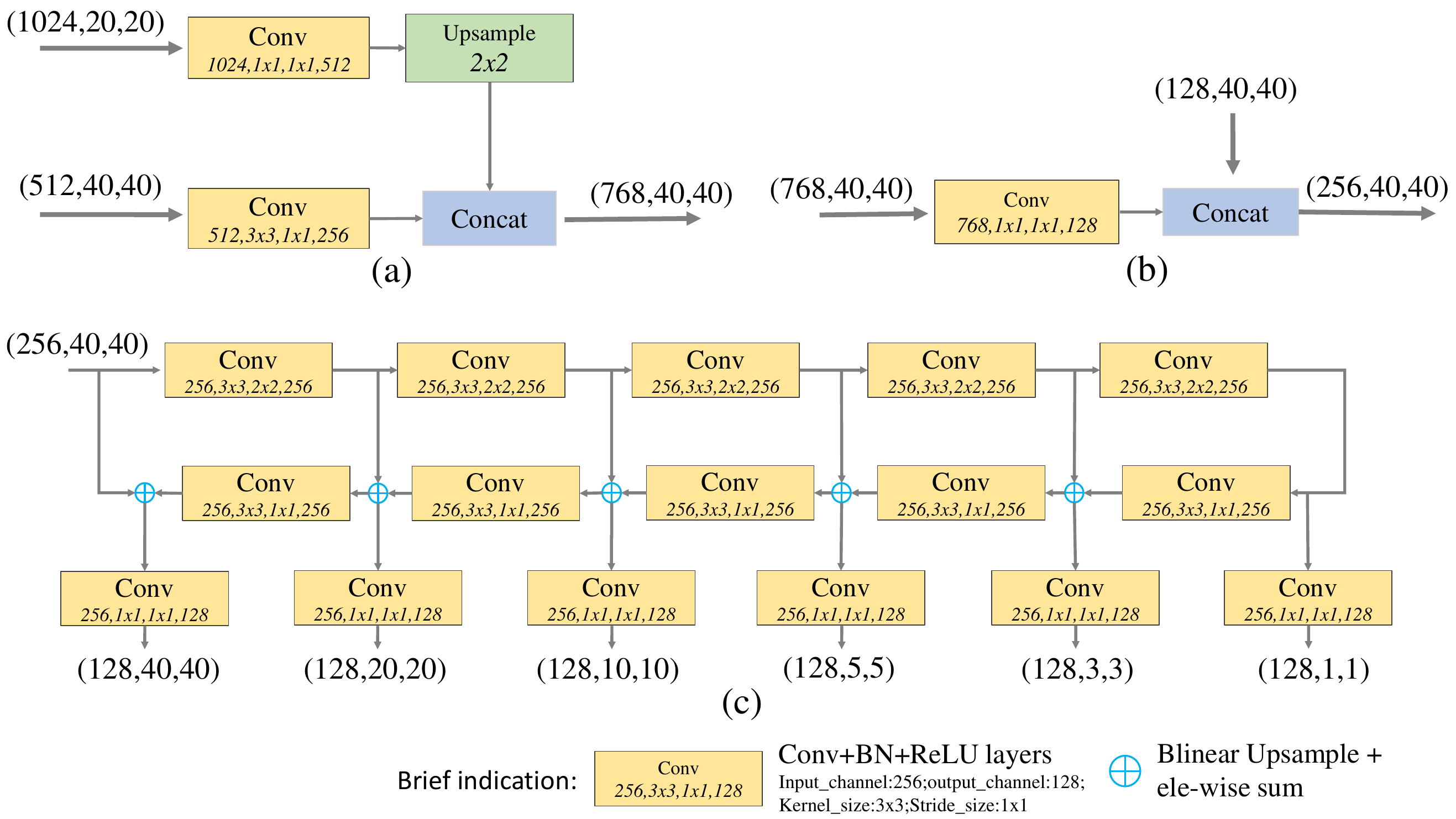}
	\caption{Structural details of some modules. (a) FFMv1, (b) FFMv2, (c) TUM. The inside numbers of each block denote: input channels, Conv kernel size, stride size, output channels.}
	\label{fig:ffb}
\end{figure}

\subsection{Multi-level Feature Pyramid Network}
As shown in Fig. \ref{fig:pipeline}, MLFPN contains three parts. Firstly, FFMv1 fuses shallow and deep features to produce the base feature, \textit{e.g.}, conv4\_3 and conv5\_3 of VGG \cite{SimonyanZ14a}, which provide multi-level semantic information for MLFPN. Secondly, several TUMs and FFMv2 are stacked alternately. Specifically, each TUM generates several feature maps with different scales. The FFMv2 fuses the base feature and the largest output feature map of the previous TUM. And the fused feature maps are fed to the next TUM. Note that the first TUM has no prior knowledge of any other TUMs, so it only learns from $\mathbf{X}_{base}$. The output multi-level multi-scale features are calculated as:
\begin{equation}
[\mathbf{x}_1^l,\mathbf{x}_2^l,...,\mathbf{x}_i^l]=
\left\{
    \begin{matrix}
             \mathbf{T}_l(\mathbf{X}_{base}), & l=1 \\
             \mathbf{T}_l(\mathbf{F}(\mathbf{X}_{base}, \mathbf{x}_i^{l-1})), & l=2...L
    \end{matrix}
\right.,
\end{equation}
where $\mathbf{X}_{base}$ denotes the base feature, $x_i^l$ denotes the feature with the $i$-th scale in the $l$-th TUM, $L$ denotes the number of TUMs, $\mathbf{T}_l$ denotes the $l$-th TUM processing, and $\textbf{F}$ denotes FFMv1 processing. Thirdly, SFAM aggregates the multi-level multi-scale features by a scale-wise feature concatenation operation and a channel-wise attention mechanism.

\textbf{FFMs} FFMs fuse features from different levels in M2Det, which are crucial to constructing the final multi-level feature pyramid. They use 1x1 convolution layers to compress the channels of the input features and use concatenation operation to aggregate these feature maps. Especially, since FFMv1 takes two feature maps with different scales in backbone as input, it adopts one upsample operation to rescale the deep features to the same scale before the concatenation operation. Meanwhile, FFMv2 takes the base feature and the largest output feature map of the previous TUM -- these two are of the same scale -- as input, and produces the fused feature for the next TUM. Structural details of FFMv1 and FFMv2 are shown in Fig. \ref{fig:ffb} (a) and (b), respectively.

\textbf{TUMs} Different from FPN \cite{LinDGHHB17} and RetinaNet \cite{LinGGHD17}, TUM adopts a thinner U-shape structure as illustrated in Fig. \ref{fig:ffb} (c). The encoder is a series of 3x3 convolution layers with stride 2. And the decoder takes the outputs of these layers as its reference set of feature maps, while the original FPN chooses the output of the last layer of each stage in ResNet backbone. In addition, we add 1x1 convolution layers after upsample and element-wise sum operation at the decoder branch to enhance learning ability and keep smoothness for the features \cite{LinCY13}. All of the outputs in the decoder of each TUM form the multi-scale features of the current level. As a whole, the outputs of stacked TUMs form the multi-level multi-scale features, while the front TUM mainly provides shallow-level features, the middle TUM provides medium-level features, and the back TUM provides deep-level features.

\textbf{SFAM}
SFAM aims to aggregate the multi-level multi-scale features generated by TUMs into a multi-level feature pyramid as shown in Fig.  \ref{fig:attention}. The first stage of SFAM is to concatenate features of the equivalent scale together along the channel dimension. The aggregated feature pyramid can be presented as $\mathbf{X} =[\mathbf{X}_1,\mathbf{X}_2,...,\mathbf{X}_i]$, where $\mathbf{X}_i = Concat(\mathbf{x}_i^1,\mathbf{x}_i^2,...,\mathbf{x}_i^L) \in \mathbb{R}^{W_{i}\times H_{i}\times C}$ refers to the features of the $i$-th largest scale. Here, each scale in the aggregated pyramid contains features from multi-level depths. However, simple concatenation operations are not adaptive enough. In the second stage, we introduce a channel-wise attention module to encourage features to focus on channels that they benefit most. Following SE block \cite{abs-1709-01507}, we use global average pooling to generate channel-wise statistics $\mathbf{z} \in \mathbb{R}^C$ at the squeeze step. And to fully capture channel-wise dependencies, the following excitation step learns the attention mechanism via two fully connected layers:
\begin{equation}
\mathbf{s} = \mathbf{F}_{ex}(\mathbf{z},\mathbf{W}) = \sigma(\mathbf{W}_{2} \delta(\mathbf{W}_{1}\mathbf{z})),
\end{equation}
where $\sigma$ refers to the ReLU function, $\delta$ refers to the sigmoid function, $\mathbf{W}_{1} \in \mathbb{R}^{\frac{C}{r}\times C}$ , $\mathbf{W}_{2} \in \mathbb{R}^{C\times \frac{C}{r}}$, r is the reduction ratio ($r=16$ in our experiments). The final output is obtained by reweighting the input $\mathbf{X}$ with activation $\mathbf{s}$:
\begin{equation}
\tilde{\mathbf{X}}_i^c = \mathbf{F}_{scale}(\mathbf{X}_i^c,s_c) = s_c \cdot \mathbf{X}_i^c,
\end{equation}
where $\tilde{\mathbf{X}_i} = [\tilde{\mathbf{X}}_i^1,\tilde{\mathbf{X}}_i^2,...,\tilde{\mathbf{X}}_i^C]$, each of the features is enhanced or weakened by the rescaling operation.

\subsection{Network Configurations}
We assemble M2Det with two kinds of backbones \cite{SimonyanZ14a,HeZRS16}. Before training the whole network, the backbones need to be pre-trained on the ImageNet 2012 dataset \cite{RussakovskyDSKS15}. All of the default configurations of MLFPN contain 8 TUMs, each TUM has 5 striding-Convs and 5 Upsample operations, so it will output features with 6 scales. To reduce the number of parameters, we only allocate 256 channels to each scale of their TUM features, so that the network could be easy to train on GPUs. As for input size, we follow the original SSD, RefineDet and RetinaNet, \textit{i.e.,}  320, 512 and 800.

At the detection stage, we add two convolution layers to each of the 6 pyramidal features to achieve location regression and classification respectively. The detection scale ranges of the default boxes of the six feature maps follow the setting of the original SSD. And when input size is 800$\times$800, the scale ranges increase proportionally except keeping the minimum size of the largest feature map. At each pixel of the pyramidal features, we set six anchors with three ratios entirely. Afterward, we use a probability score of 0.05 as threshold to filter out most anchors with low scores. Then we use soft-NMS \cite{BodlaSCD17} with a linear kernel for post-processing, leaving more accurate boxes. Decreasing the threshold to 0.01 can generate better detection results, but it will slow down the inference time a lot, we do not consider it for pursuing better practical values.

\section{Experiments}
In this section, we present experimental results on the bounding box detection task of the challenging MS-COCO benchmark. Following the protocol in MS-COCO, we use the \texttt{trainval35k} set for training, which is a union of 80k images from \texttt{train} split and a random 35 subset of images from the 40k image \texttt{val} split. To compare with state-of-the-art methods, we report COCO AP on the \texttt{test-dev} split, which has no public labels and requires the use of the evaluation server. And then, we report the results of ablation studies evaluated on the \texttt{minival} split for convenience.

Our experiment section includes 4 parts: (1) introducing implement details about the experiments; (2) demonstrating the comparisons with state-of-the-art approaches; (3) ablation studies about M2Det; (4) comparing different settings about the internal structure of MLFPN and introducing several version of M2Det.

\begin{table*}[!t]
\centering
 \scriptsize
	\caption{Detection accuracy comparisons in terms of mAP percentage on \textit{MS COCO} test-dev set.}
	\label{tab:cocoresults}
	\centering
	\begin{tabular}{|p{5cm}<{\raggedright}|p{1.4cm}<{\raggedright}|p{1.3cm}<{\centering}|p{1.0cm}<{\centering}|p{0.8cm}<{\centering}|p{0.6cm}<{\centering}p{0.6cm}<{\centering}p{0.6cm}<{\centering}|p{0.6cm}<{\centering}p{0.6cm}<{\centering}p{0.6cm}<{\centering}|}
		\hline
		\multirow{2}{*}{Method} & \multirow{2}{*}{Backbone} & \multirow{2}{*}{Input size} &  \multirow{2}{*}{MultiScale} & \multirow{2}{*}{FPS} &\multicolumn{3}{c|}{Avg. Precision, IoU:} & \multicolumn{3}{c|}{Avg. Precision, Area:}  \\
		& & & & & 0.5:0.95 & 0.5 & 0.75 & S & M & L \\
		\hline
		\small\textit{two-stage:} & & & & & & & & & &\quad \\
		Faster R-CNN \cite{RenHGS15}   & VGG-16 & $\sim$1000$\times$600&False&7.0& 21.9 & 42.7 & - & - & - & - \\
		OHEM++ \cite{ShrivastavaGG16}  & VGG-16 & $\sim$1000$\times$600&False&7.0& 25.5 & 45.9 & 26.1 & 7.4 & 27.7 & 40.3 \\
		R-FCN \cite{DaiLHS16} & ResNet-101 & $\sim$1000$\times$600&False&9& 29.9 & 51.9 & - & 10.8 & 32.8 & 45.0 \\
		CoupleNet \cite{ZhuZWZWL17}  & ResNet-101 & $\sim$1000$\times$600&False&8.2& 34.4 & 54.8 & 37.2 & 13.4 & 38.1 & 50.8 \\
		Faster R-CNN w FPN \cite{LinDGHHB17}  & Res101-FPN & $\sim$1000$\times$600&False&6& 36.2 & 59.1 & 39.0 & 18.2 & 39.0 & 48.2 \\
		Deformable R-FCN \cite{DaiQXLZHW17}  & Inc-Res-v2 & $\sim$1000$\times$600&False&-& 37.5 & 58.0 & 40.8 & 19.4 & 40.1 & 52.5 \\
		Mask R-CNN \cite{HeGDG17} &ResNeXt-101 &$\sim$1280$\times$800 & False &3.3& 39.8 & 62.3 & 43.4 & 22.1 & 43.2& 51.2\\
		Fitness-NMS \cite{abs-1711-00164} &ResNet-101 & $\sim$1024$\times$1024 &True & 5.0 & 41.8&60.9 &44.9& 21.5&45.0&57.5\\
		Cascade R-CNN \cite{abs-1712-00726} &Res101-FPN &$\sim$1280$\times$800&False&7.1& 42.8 &62.1&46.3&23.7&45.5&55.2\\
		SNIP \cite{abs-1711-08189} &DPN-98 & - & True & - & 45.7 & 67.3 & 51.1 & 29.3 & 48.8 & 57.1 \\
		\hline
    		\hline
		\small\textit{one-stage:} & & & & & & & & & &\quad\\
		SSD300* \cite{LiuAESRFB16}  & VGG-16 & 300$\times$300&False&43& 25.1 & 43.1 & 25.8 & 6.6 & 25.9 & 41.4 \\
		RON384++ \cite{KongSYLLC17}  & VGG-16 & 384$\times$384&False&15& 27.4 & 49.5 & 27.1 & - & - & - \\
		DSSD321 \cite{FuLRTB17}  & ResNet-101 &321$\times$321&False&9.5& 28.0 & 46.1 & 29.2 & 7.4 & 28.1 & 47.6 \\
		RetinaNet400 \cite{LinGGHD17} &ResNet-101 & $\sim$640$\times$400 &False&12.3& 31.9 & 49.5& 34.1 & 11.6 & 35.8 & 48.5\\
		RefineDet320 \cite{abs-1711-06897}  & VGG-16 & 320$\times$320&False&38.7& 29.4 & 49.2 & 31.3 & 10.0 & 32.0 & 44.4 \\
		RefineDet320 \cite{abs-1711-06897}  & ResNet-101 & 320$\times$320&True&-&38.6& 59.9 & 41.7 & 21.1 & 41.7 & 52.3 \\

		\textbf{M2Det (Ours)}& VGG-16& 320$\times$320&False&33.4&33.5&52.4&35.6&14.4&37.6&47.6\\
		\textbf{M2Det (Ours)}& VGG-16& 320$\times$320&True&-&38.9&59.1&42.4&24.4&41.5&47.6\\
		\textbf{M2Det (Ours)}& ResNet-101&320$\times$320&False&21.7&34.3&53.5&36.5&14.8&38.8&47.9\\
		\textbf{M2Det (Ours)}& ResNet-101&320$\times$320&True&-&\textbf{39.7}&\textbf{60.0}&\textbf{43.3}&\textbf{25.3}&\textbf{42.5}&\textbf{48.3}\\

		\hline
		YOLOv3 \cite{yolov3} & DarkNet-53 & 608$\times$608 &False&19.8& 33.0 & 57.9 & 34.4 & 18.3 & 35.4 & 41.9 \\
		SSD512* \cite{LiuAESRFB16}  & VGG-16 & 512$\times$512&False&22& 28.8 & 48.5 & 30.3 & 10.9 & 31.8 & 43.5 \\
		DSSD513 \cite{FuLRTB17}  & ResNet-101 & 513$\times$513&False&5.5& 33.2 & 53.3 & 35.2 & 13.0 & 35.4 & 51.1 \\
		RetinaNet500 \cite{LinGGHD17}  & ResNet-101 & $\sim$832$\times$500 &False&11.1& 34.4 & 53.1 & 36.8 & 14.7 & 38.5 & 49.1 \\
		RefineDet512 \cite{abs-1711-06897} & VGG-16 &512$\times$512&False&22.3& 33.0 & 54.5 & 35.5 & 16.3 & 36.3 & 44.3 \\
		RefineDet512 \cite{abs-1711-06897} & ResNet-101 &512$\times$512&True&-& 41.8 & 62.9 & 45.7 & 25.6 & 45.1 & 54.1 \\
		CornerNet \cite{abs-1808-01244} & Hourglass & 512$\times$512&False&4.4& 40.5&57.8&45.3&20.8&44.8&56.7\\
		CornerNet \cite{abs-1808-01244} & Hourglass & 512$\times$512&True&-& 42.1&57.8&45.3&20.8&44.8&56.7\\
		\textbf{M2Det (Ours)}& VGG-16&512$\times$512&False&18.0&37.6&56.6&40.5&18.4&43.4&51.2\\
		\textbf{M2Det (Ours)}& VGG-16&512$\times$512&True&-& 42.9 & 62.5&47.2&28.0&47.4&52.8\\
		\textbf{M2Det (Ours)}& ResNet-101&512$\times$512&False&15.8&38.8&59.4&41.7&20.5&43.9&53.4\\
		\textbf{M2Det (Ours)}& ResNet-101&512$\times$512&True&-&\textbf{43.9}&\textbf{64.4}&\textbf{48.0}&\textbf{29.6}&\textbf{49.6}&\textbf{54.3}\\

	    \hline
	    RetinaNet800 \cite{LinGGHD17}  & Res101-FPN & $\sim$1280$\times$800 &False&5.0& 39.1 & 59.1 & 42.3 & 21.8 & 42.7 & 50.2 \\

		\textbf{M2Det (Ours)}& VGG-16&800$\times$800&False&11.8&41.0&59.7&45.0&22.1&46.5&53.8\\
		\textbf{M2Det (Ours)}& VGG-16&800$\times$800&True&-&\textbf{44.2}&\textbf{64.6}&\textbf{49.3}&\textbf{29.2}&\textbf{47.9}&\textbf{55.1}\\
		\bottomrule
	\end{tabular}
	\label{tab:state-of-th-art}
\end{table*}

\subsection{Implementation details}
For all experiments based on M2Det, we start training with warm-up strategy for 5 epochs, initialize the learning rate as $2 \times 10^{-3}$, and then decrease it to $2 \times 10^{-4}$ and $2 \times 10^{-5}$ at 90 epochs and 120 epochs, and stop at 150 epochs. M2Det is developed with PyTorch v0.4.0 \footnote{\url{https://pytorch.org/}}. When input size is 320 and 512, we conduct experiments on a machine with 4 NVIDIA Titan X GPUs, CUDA 9.2 and cuDNN 7.1.4, while for input size of 800, we train the network on NVIDIA Tesla V100 to get results faster. The batch size is set to 32 (16 each for 2 GPUs, or 8 each for 4 GPUs). On NVIDIA Titan Xp that has 12 GB memory, the training performance is limited if batch size on a single GPU is less than 5. Notably, for Resnet101, M2Det with the input size of 512 is not only limited in the batch size (only 4 is available), but also takes a long time to train, so we train it on V100.

For training M2Det with the VGG-16 backbone when input size is 320$\times$320 and 512$\times$512 on 4 Titan X devices, the total training time costs 3 and 6 days respectively, and with the ResNet-101 backbone when 320$\times$320 costs 5 days. While for training M2Det with ResNet-101 when input size is 512$\times$512 on 2 V100 devices, it costs 11 days. The most accurate model is M2Det with the VGG backbone and 800$\times$800 input size, it costs 14 days.

\subsection{Comparison with State-of-the-art}
We compare the experimental results of the proposed M2Det with state-of-the-art detectors in Table \ref{tab:state-of-th-art}. For these experiments, we use 8 TUMs and set 256 channels for each TUM. The main information involved in the comparison includes the input size of the model, the test method (whether it uses multi-scale strategy), the speed of the model, and the test results. Test results of M2Det with 10 different setting versions are reported in Table \ref{tab:state-of-th-art}, which are produced by testing it on MS-COCO \texttt{test-dev} split, with a single NVIDIA Titan X PASCAL and the batch size 1. Other statistical results stem from references. It is noteworthy that, M2Det-320 with VGG backbone achieves AP of 38.9, which has surpassed most object detectors with more powerful backbones and larger input size, \textit{e.g.}, AP of Deformable R-FCN \cite{DaiQXLZHW17} is 37.5, AP of Faster R-CNN with FPN is 36.2. Assembled with ResNet-101 can further improve M2Det, the single-scale version obtains AP of 38.8, which is competitive with state-of-the-art two-stage detectors Mask R-CNN \cite{HeGDG17}. In addition, based on the optimization of PyTorch, it can run at 15.8 FPS. RefineDet \cite{abs-1711-06897} inherits the merits of one-stage detectors and two-stage detectors, gets AP of 41.8, CornerNet \cite{abs-1808-01244} proposes key point regression for detection and borrows the advantages of Hourglass \cite{NewellYD16} and focal loss \cite{LinGGHD17}, thus gets AP of 42.1. In contrast, our proposed M2Det is based on the regression method of original SSD, with the assistance of Multi-scale Multi-level features, obtains 44.2 AP, which exceeds all one-stage detectors. Most approaches do not compare the speed of multi-scale inference strategy due to different methods or tools used, so we also only focus on the speed of single-scale inference methods.

In addition, in order to emphasize that the improvement of M2Det is not entirely caused by the deepened depth of the model or the gained parameters, we compare with state-of-the-art one-stage detectors and two-stage detectors. CornerNet with Hourglass has 201M parameters, Mask R-CNN \cite{HeGDG17} with ResNeXt-101-32x8d-FPN \cite{XieGDTH17} has 205M parameters. By contrast, M2Det800-VGG has only 147M parameters. Besides, consider the comparison of depth, it is also not dominant.

\subsection{Ablation study}
Since M2Det is composed of multiple subcomponents, we need to verify each of its effectiveness to the final performance. The baseline is a simple detector based on the original SSD, with 320$\times$320 input size and VGG-16 reduced backbone.

\begin{table}[h]
\centering
\scriptsize
\caption{Ablation study of M2Det. The detection results are evaluated on \texttt{minival} set}
\label{tab:SurveyDetails}
\begin{tabular}{l||ccccccc}
\toprule
+ 1 s-TUM & {} &${\checkmark}$& {}  &{} &{}& {}  \\
+ 8 s-TUM & {} &{}&${\checkmark}$ &{} &{}& {}   \\
+ 8 TUM &{}&{} & {} & ${\checkmark}$ & ${\checkmark}$& ${\checkmark}$ & ${\checkmark}$   \\
+ Base feature &{} & {} &  {}&{} & ${\checkmark}$ & ${\checkmark}$ & ${\checkmark}$ \\
+ SFAM & {}& {} &{} & {} & {} & ${\checkmark}$ & ${\checkmark}$\\
VGG16 $\Rightarrow$ Res101 & {}&{} &{} &  {} & {} & {} & ${\checkmark}$ \\

\midrule
AP & 25.8& 27.5&30.6 & 30.8 & 32.7 & 33.2  &\textbf{34.1}\\ 
AP$_{50}$ & 44.7& 45.2&50.0 & 50.3 & 51.9 & 52.2 &\textbf{53.7}\\ 
AP$_{small}$ & 7.2& 7.7 &13.8 &13.7 & 13.9 & 15.0 &\textbf{15.9}\\ 
AP$_{medium}$ & 27.4& 28.0 &35.3& 35.3 & 37.9 & 38.2 &\textbf{39.5}  \\ 
AP$_{large}$ & 41.4& 47.0 & 44.5&44.8 & 48.8 & 49.1  &\textbf{49.3}\\
\bottomrule
\end{tabular}
\label{tab:Margin_settings}
\end{table}

\textbf{TUM} To demonstrate the effectiveness of TUM, we conduct three experiments. First, following DSSD, we extend the baseline detector with a series of Deconv layers, and the AP has improved from 25.8 to 27.5 as illustrated in the third column in Table \ref{tab:Margin_settings}. Then we replace with MLFPN into it. As for the U-shape module, we firstly stack 8 s-TUMs, which is modified to decrease the 1$\times$1 Convolution layers shown in Fig. \ref{fig:ffb}, then the performance has improved 3.1 compared with the last operation, shown in the forth column in Table \ref{tab:Margin_settings}. Finally, replacing TUM by s-TUM in the fifth column has reached the best performance, it comes to AP of 30.8.

\textbf{Base feature} Although stacking TUMs can improve detection, but it is limited by input channels of the first TUM. That is, decreasing the channels will drop the abstraction of MLFPN, while increasing them will highly increase the parameters number. Instead of using base feature only once, We afferent base feature at the input of each TUM to alleviate the problem. For each TUM, the embedded base feature provides necessary localization information since it contains shallow features. The AP percentage increases to 32.7, as shown in the sixth column in Table \ref{tab:Margin_settings}.

\textbf{SFAM}
 As shown in the seventh column in Table \ref{tab:Margin_settings}, compared with the architecture that without SFAM, all evaluation metrics have been upgraded. Specifically, all boxes including small, medium and large become more accurate.

\textbf{Backbone feature}
As in many visual tasks, we observe a noticeable AP gain from $33.2$ to $34.1$ when we use well-tested ResNet-101 \cite{HeZRS16} instead of VGG-16 as the backbone network. As shown in Table \ref{tab:Margin_settings}, such observation remains true and consistent with other AP metrics.

\begin{table}[h]
\scriptsize
\centering
\begin{tabular}{cc||cccc}
\toprule
TUMs &Channels&Params(M)&AP&AP$_{50}$&AP$_{75}$\\
\hline\hline
2& 256 & 40.1 & 30.5 &50.5&32.0\\
2& 512 & 106.5& 32.1&51.8&34.0\\
4& 128 &34.2& 29.8& 49.7&31.2\\
4& 256 &60.2& 31.8&51.4&33.0\\
4& 512 &192.2 & 33.4 & 52.6 & 34.2\\
8& 128 &47.5& 31.8& 50.6&33.6\\
8& 256 &98.9& 33.2& 52.2&35.2\\
8& 512 &368.8& 34.0& 52.9&36.4\\
16& 128 &73.9 &32.5 &51.7&34.4\\
16& 256 & 176.8&33.6 &52.6&35.7\\
\bottomrule
\end{tabular}
\caption{Different configurations of MLFPN in M2Det. The backbone is VGG and input image is 320$\times$320.}
\label{tab:tum_ablation}
\end{table}

\subsection{Variants of MLFPN}
The Multi-scale Multi-level Features have been proved to be effective. But what is the boundary of the improvement brought by MLFPN? Step forward, how to design TUM and how many TUMs should be OK? We implement a group of variants to find the regular patterns. To be more specific, we fix the backbone as VGG-16 and the input image size as 320x320, and then tune the number of TUMs and the number of internal channels of each TUM.

As shown in Table \ref{tab:tum_ablation}, M2Det with different configurations of TUMs is evaluated on COCO \texttt{minival} set. Comparing the number of TUMs when fixing the channels, \textit{e.g.},256, it can be concluded that stacking more TUMs brings more promotion in terms of detection accuracy. Then fixing the number of TUMs, no matter how many TUMs are assembled, more channels consistently benefit the results. Furthermore, assuming that we take a version with 2 TUMS and 128 channels as the baseline, using more TUMs could bring larger improvement compared with increasing the internal channels, while the increase in parameters remains similar.

\subsection{Speed}
\begin{figure}[t]
\centering
\includegraphics[scale=0.35]{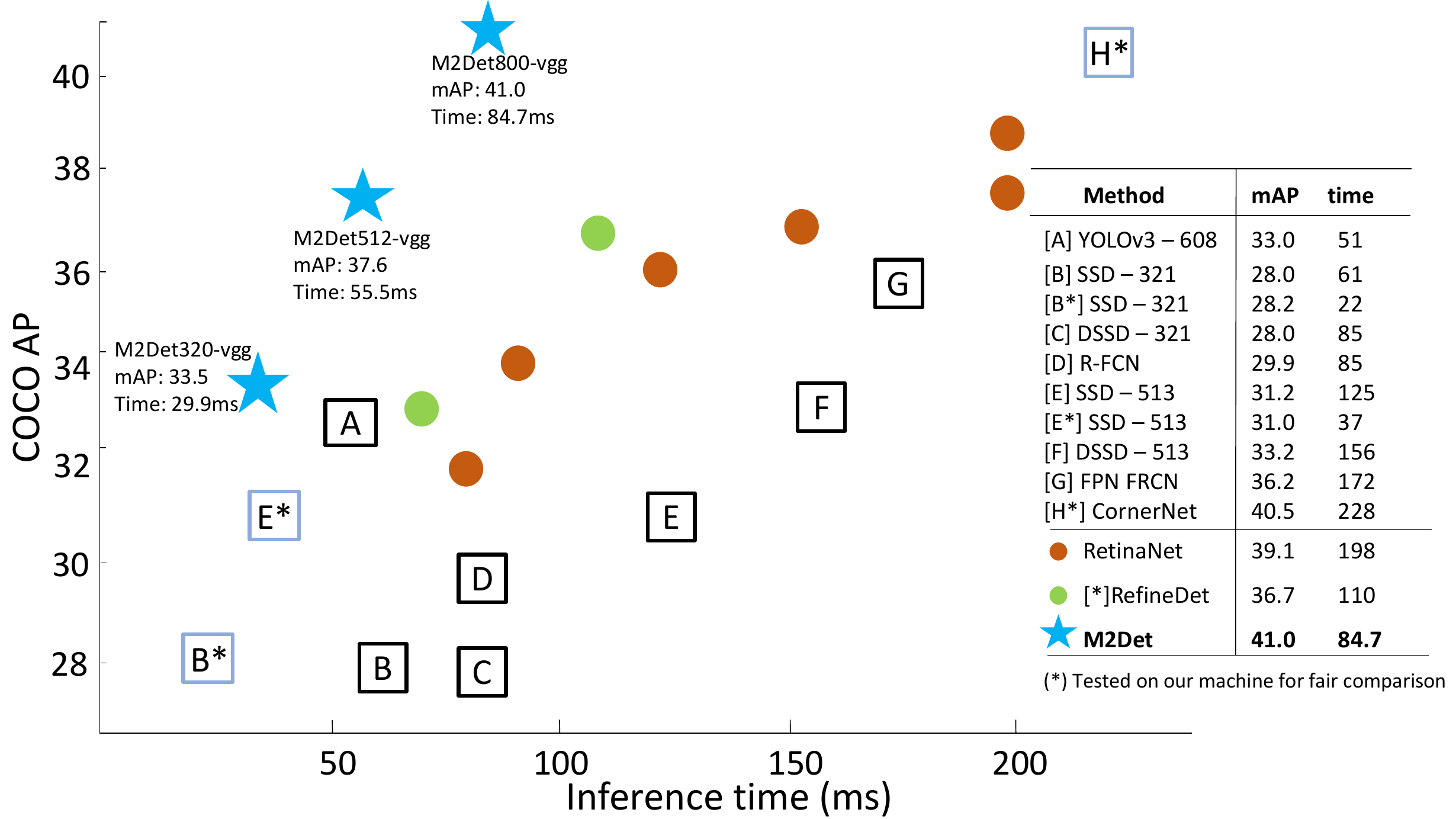}
\caption{Speed (ms) vs. accuracy (mAP) on COCO \textit{test-dev}.}
\label{fig:rank}
\end{figure}

We compare the inference speed of M2Det with state-of-the-art approaches. Since VGG-16 \cite{SimonyanZ14a} reduced backbone has removed FC layers, it is very fast to use it for extracting base feature. We set the batch size to 1, take the sum of the CNN time and NMS time of 1000 images, and divide by 1000 to get the inference time of a single image. Specifically, we assemble VGG16-reduced to M2Det and propose the fast version M2Det with the input size 320$\times$320, the standard version M2Det with 512$\times$512 input size and the most accurate version M2Det with 800$\times$800 input size. Based on the optimization of PyTorch, M2Det can achieve accurate results with high speed. As shown in Fig. \ref{fig:rank}, M2Det benefits the advantage of one-stage detection and our proposed MLFPN structure, draws a significantly better speed-accuracy curve compared with other methods. For fair comparison, we reproduce and test the speed of SSD321-ResNet101, SSD513-ResNet101 \cite{FuLRTB17}, RefineDet512-ResNet101, RefineDet320-ResNet101 \cite{abs-1711-06897} and CornerNet \cite{abs-1808-01244} on our device. It is clear that M2Det performs more accurately and efficiently.

\section{Discussion}
We think the detection accuracy improvement of M2Det is mainly brought by the proposed MLFPN. On one hand, we fuse multi-level features extracted by backbone as the base feature, and then feed it into a block of alternating joint Thinned U-shape Modules and Feature Fusion Modules to extract more representative, multi-level multi-scale features, \textit{i.e.} the decoder layers of each TUM. Obviously, these decoder layers are much deeper than the layers in the backbone, and thus more representative for object detection. Contrasted with our method, the existing detectors \cite{abs-1711-06897,LinDGHHB17,FuLRTB17} just use the layers of the backbone or extra layers with few depth increase. Hence, our method can achieve superior detection performance. On the other hand, each feature map of the multi-level feature pyramid generated by the SFAM consists of the decoder layers from multiple levels. In particular, at each scale, we use multi-level features to detect objects, which would be better for \textbf{handling appearance-complexity variation across object instances}.

\begin{figure}[t]
\centering
\includegraphics[scale=0.4]{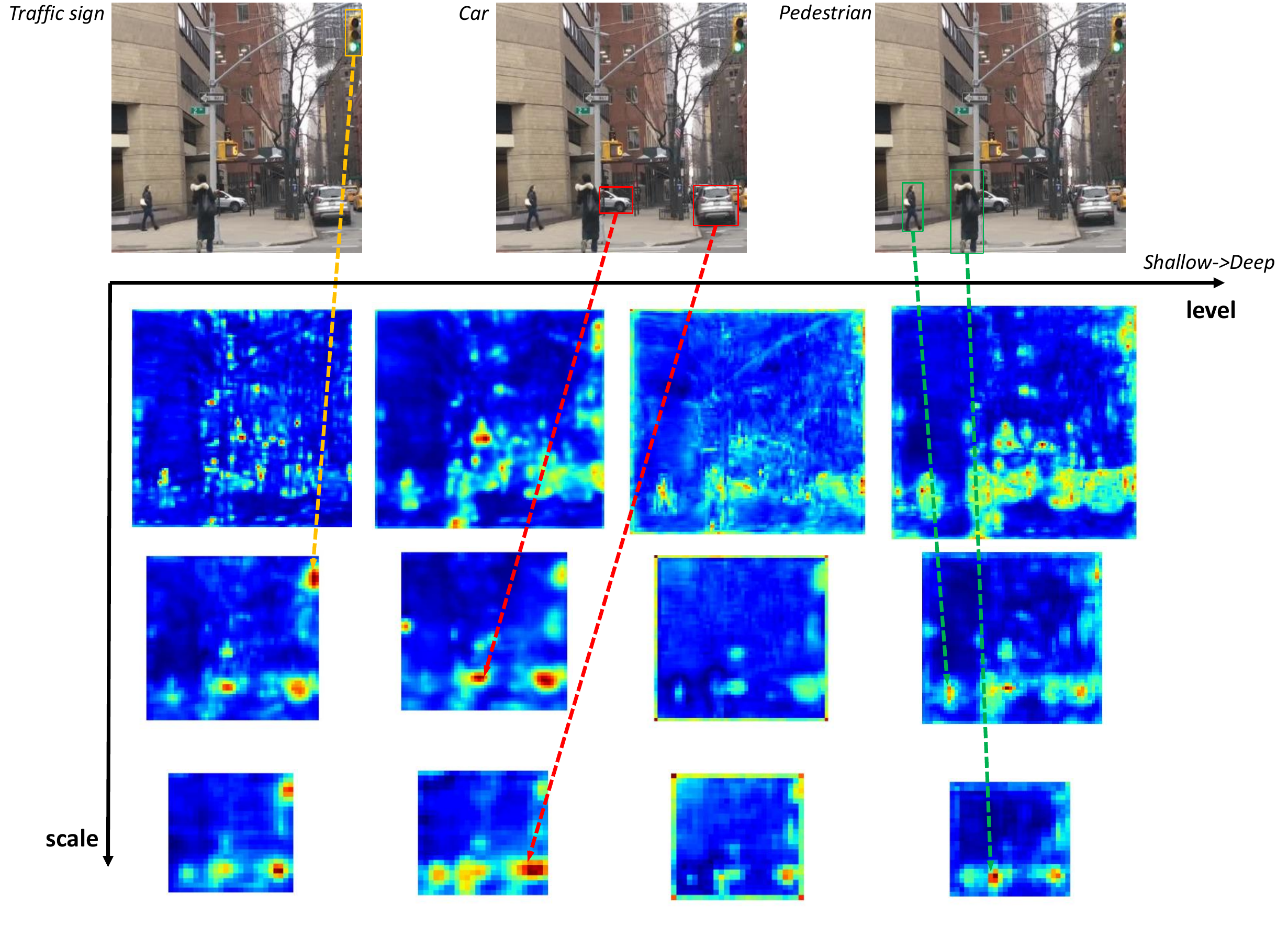}
\caption{Example activation values of multi-scale multi-level features. Best view in color.}
\label{fig:vis}
\end{figure}
To verify that the proposed MLFPN can learn effective feature for detecting objects with different scales and large appearance variation, we visualize the activation values of classification Conv layers along scale and level dimensions, such an example shown in Fig. \ref{fig:vis}. The input image contains two persons, two cars and a traffic light. Moreover, the sizes of the two persons are different, as well as the two cars. And the traffic light, the smaller person and the smaller car have similar sizes. We can find that: 1) compared with the smaller person, the larger person has strongest activation value at the feature map of large scale, so as to the smaller car and larger car; 2) the traffic light, the smaller person and the smaller car have strongest activation value at the feature maps of the same scale; 3) the persons, the cars and the traffic light have strongest activation value at the highest-level, middle-level, lowest-level feature maps respectively. This example presents that: 1) our method learns very effective features to handle scale variation and appearance-complexity variation across object instances; 2) it is necessary to use multi-level features to detect objects with similar size.

\section{Conclusion}
In this work, a novel method called Multi-Level Feature Pyramid Network (MLFPN) is proposed to construct effective feature pyramids for detecting objects of different scales. MLFPN consists of several novel modules. First, multi-level features (\textit{i.e.} multiple layers) extracted by backbone are fused by a Feature Fusion Module (FFMv1) as the base feature. Second, the base feature is fed into a block of alternating joint Thinned U-shape Modules (TUMs) and Fature Fusion Modules (FFMv2s) and multi-level multi-scale features (\textit{i.e.} the decoder layers of each TUM) are extracted. Finally, the extracted multi-level multi-scale features with the same scale (size) are aggregated to construct a feature pyramid for object detection by a Scale-wise Feature Aggregation Module (SFAM). A powerful end-to-end one-stage object detector called M2Det is designed based on the proposed MLFPN, which achieves a new state-of-the-art result (\textit{i.e.} AP of 41.0 at speed of 11.8 FPS with single-scale inference strategy and AP of 44.2 with multi-scale inference strategy) among the one-stage detectors on MS-COCO benchmark. Additional ablation studies further demonstrate the effectiveness of the proposed architecture and the novel modules.

\section{Acknowledgements}
This work is supported by National Natural Science Foundation of China under Grant
61673029. This work is also a research achievement of Key Laboratory of Science, Technology and Standard
in Press Industry (Key Laboratory of Intelligent Press Media Technology).

\bibliographystyle{aaai}
\bibliography{m2det}
\end{document}